\documentclass[12pt, a4paper, twocolumn]{article}

\usepackage[colorlinks,urlcolor=blue,linkcolor=blue,citecolor=blue]{hyperref}
\usepackage{amsmath}
\usepackage{authblk}
\usepackage{booktabs}
\usepackage{comment}
\usepackage{float} 
\usepackage{graphicx}
\usepackage{lettrine}
\usepackage{multicol}

\expandafter\def\expandafter\UrlBreaks\expandafter{\UrlBreaks\do\/\do\*\do\-\do\~\do\'\do\"\do\-}

\begin{document}

\title{Results of Low Power Computer Vision Challenge (LPCVC) 2023}

\author{Leo Chen and Benjamin Boardley}
\affil{Purdue University, West Lafayette, IN, 47906, USA}

\author{Ping Hu}
\affil{Boston University, Boston, MA, 02215, UAS}

\author{Yiru Wang, Yifan Pu, Xin Jin, Yongqiang Yao, Ruihao Gong, Bo Li, Gao Huang, Xianglong Liu}
\affil{ModelTC and Tsinghua University, Beijing, China}

\author{Zifu Wan, Xinwang Chen, Ning Liu, Ziyi Zhang, Dongping Liu, Ruijie Shan, Zhengping Che, Fachao Zhang, Xiaofeng Mou, Jian Tang}
\affil{Midea Group, Beijing, China}

\author{Maxim Chuprov, Ivan Malofeev, Alexander Goncharenko, Andrey Shcherbin, Arseny Yanchenko, Sergey Alyamkin}
\affil{enot.ai, Grand Duchy of Luxembourg, Luxembourg}

\author{Xiao Hu}
\affil{Qualcomm, San Diego, CA, 92121, USA}

\author{George K. Thiruvathukal}
\affil{{Loyola} University Chicago, Chicago, IL, 60660, USA}

\author{Yung Hsiang Lu}
\affil{Purdue University, West Lafayette, IN, 47906, USA}
\markboth{THEME/FEATURE/DEPARTMENT}{THEME/FEATURE/DEPARTMENT}

\twocolumn[
  \begin{@twocolumnfalse}
    \maketitle
    \begin{abstract}
    This article describes the 2023 IEEE Low-Power Computer Vision Challenge (LPCVC). 
Since 2015, LPCVC has been an international competition devoted to tackling the challenge of computer vision (CV) on edge devices. 
Most CV researchers focus on improving accuracy, at the expense of ever-growing sizes of machine models. LPCVC balances accuracy with resource requirements.  
Winners must achieve high accuracy with short execution time when their CV solutions run on an embedded device, such as Raspberry PI or  Nvidia Jetson Nano.  
The vision problem for 2023 LPCVC is segmentation of images acquired by Unmanned Aerial Vehicles (UAVs, also called drones) after disasters. 
The 2023 LPCVC attracted 60 international teams
that submitted 676 solutions during the submission window of one month.
This article explains the setup of the competition and highlights
the winners' methods that improve accuracy and shorten execution time.

    \end{abstract}
  \end{@twocolumnfalse}
]
\maketitle
Competitions have been a strong driver for innovations. The impressive progress of computer vision in the recent decade is driven partially by competitions, such as ImageNet Large Scale Visual Recognition Challenge (ILSVRC). Many computer vision (CV) competitions focus exclusively on accuracy. As a result,
the sizes of machine models (measured by the number of parameters) for CV have been increasing rapidly. Meanwhile, many CV applications must run on embedded systems with limited computing resources, for example, uncrewed aerial vehicles (UAVs, also called drones). Running since 2015, the IEEE Low-Power Computer Vision Challenge (LPCVC)~\cite{website_ieee_lpcvc} is a competition balancing accuracy and resource requirements (execution time, energy consumption, memory capacity) by running computer vision software on embedded devices (e.g., Nvidia Jetson and Raspberry PI).
The contestants submit their solutions through an online portal and the rankings are determined by the ratio of accuracy and resource usage (execution time or energy consumption). Over the years, more than 200 teams submitted more than 1,700 solutions. The history of LPCVC and winners' solutions
are available in ~\cite{21lpcvc_hu, alyamkin_low-power_2019, thiruvathukal_low-power_2022, 9458522}.

\begin{table}[h]
\centering 
\begin{tabular}{|l|r|r|} \hline
{\bf Year} & {\bf Teams} & {\bf Submissions} \\ \hline
2018 & 21 & 131 \\
2019 & 22 & 234 \\
2020 & 46 & 378 \\
2021 & 53 & 366  \\
2023 & 117 & 676 \\ \hline
{\bf Total} & 259 & 1,785 \\ 
\hline 
\end{tabular} 
\caption{Since 2018, LPCVC has
hosted 259 research  teams that submitted 1,785 solutions.}  
\label{figure:numbers_LPCVC}
\end{table}

\section{2023 LPCVC}

\begin{figure*}[ht!]
    \centering
    \includegraphics[width=1.0\textwidth]{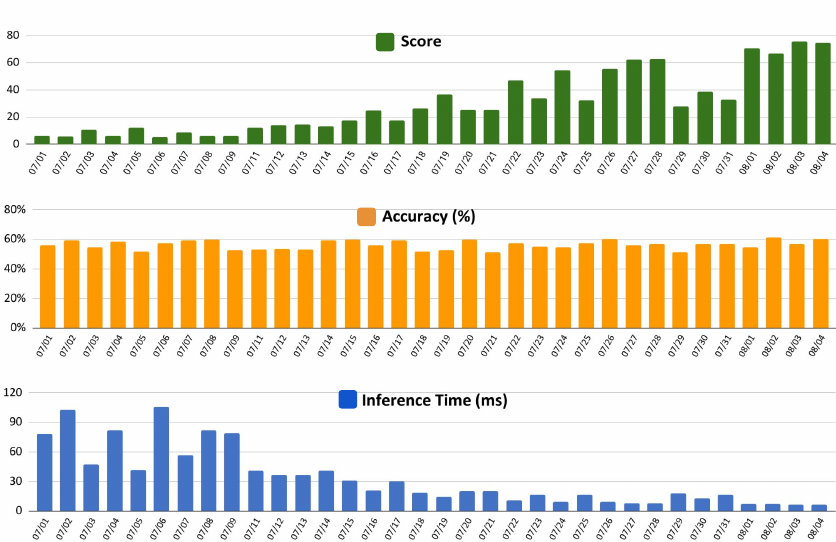}
    \caption{The highest score, highest accuracy, and lowest execution time  
    on each day during 2023 competition. The competition was open for four additional days as compensation for an outage that paused the scoring system.}
    \label{fig:progress}
\end{figure*}

\begin{figure}[h!]
    \centering
    \includegraphics[width=0.5\textwidth]{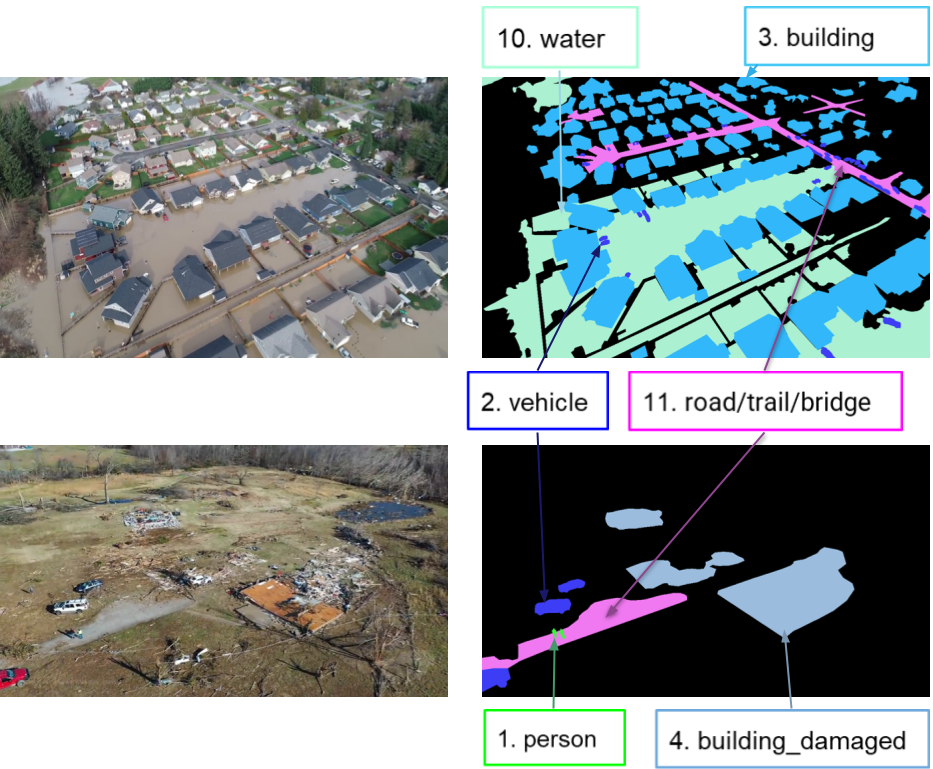}
    \caption{The 2023 challenge is semantic segmentation. This figure shows two examples of the input data.}
    \label{fig:enter-label}
\end{figure}

The 2023 LPCVC features semantic segmentation, a computer vision task where each pixel of an input image is categorized into a predetermined set of objects. Semantic segmentation can be used for disaster rescue for rapid scene assessment, identifying areas of risk and individuals to be saved~\cite{TDOT_Drones}.
Figure~\ref{fig:enter-label} shows
two examples.
Accuracy is of the essence, as an incorrect assessment may result in critical time loss. However, accuracy typically demands intense computational resources where they might not always be available. For example, drones are ideal for image capture in post-disaster scenes and their navigation algorithms could benefit from semantic segmentation. Regardless, drones must be lightweight, which limits the computational resources they may carry. Thus, the purpose of the competition is to promote the development of accurate yet efficient semantic segmentation models.

The images used are scenes captured by UAVs post-disasters. UAVs have already been utilized in several notable disaster events such as Hurricane Irma and the Mexico City Earthquake ~\cite{TDOT_Drones} to great effect. Both autonomous and operated drones were used to identify key locations to set up relief bases, create 3D mappings of the scene, and more.  

The 2023 LPCVC's evaluation test set consists of 600 images at 512 $\times$ 512  resolution. Each was hand-labeled to create ground truth for comparison. From those images, contestants are required to categorize each pixel as one of the 14 following possibilities common in disaster scenes.

\noindent
\begin{table}[h]
\fontsize{9}{9}\selectfont
\begin{tabular}{ll}
\label{table:classes}
0. background                 & 7. flood/water/river \\
1. avalanche                 & 8. ice flow        \\
2. building undamaged       & 9. lava flow            \\
3. building damaged         & 10. person               \\
4. cracks/fissures/subsidence & 11. pyroclastic flow    \\
5. debris/mud/rock flow      & 12. road/bridge  \\
6. fire/flare                & 13. vehicle     \\        
\end{tabular}
\caption{Possible labels for pixels}
\end{table}

\subsection{REFERENCE SOLUTION}

The organizers provided an open-source reference solution on GitHub~\cite{sampleSolution} as the baseline for competition results. The reference solution serves a two-fold purpose as it is used to give an example of submission format while also setting the qualification standard. A submitted solution is \textbf{disqualified} if it performs worse than the sample solution on either of the scoring metrics: accuracy or time. Our sample solution scored 50.11 in accuracy and had an average inference time of 200ms per image. The sample solution of the 2023 competition was based upon the FANet architecture~\cite{hu2020realtime}. FANet (Fast Attention Net) is an optimization of the self-attention mechanism that captures the same spatial context but reduces the computational cost. This optimization makes it ideal for a competition that focuses on low-power computer vision. 
 
\url{https://github.com/feinanshan/FANet}

\subsection{Data Set}
Sample training data (a set of 1000 images) was provided that reflected the testing data. Each image contains multiple elements from the list of possible labels. However, contestants were free to use any training data they could source. See Figure~\ref{fig:enter-label} for two examples of the data set and their corresponding ground truths.

\section{EVALUATION}

 The evaluation metrics chosen are in correspondence with the challenge of analyzing disaster scene images aboard a UAV. Solutions need to be efficient in time and energy while also giving accurate predictions. We chose the NVIDIA Jetson Nano 2GB Developer KIT, running in a power-efficient mode (5W), evaluating for model inference time and accuracy. In the following sections, we will elaborate on six components that contribute to the evaluation metrics.

\begin{enumerate}
    \item \textbf{Class Set (C)}: This parameter represents the collection of classes relevant to a particular prediction map and the correlating ground truth. We denote the set of classes in a model's prediction map as \(C_p\) and the set of classes in the ground truth as \(C_g\). The union of these sets(symbolized as C) is expressed in equation \ref{equation:Classes}.
    \begin{equation}
    \label{equation:Classes}
    C = C_p \cup C_g
    \end{equation}
    \item \textbf{True Positive (\(TP\))}: refers to the count of correctly labeled pixels for a specific class.
    \item \textbf{False Positive (\(FP\))}: indicates the number of pixels that have been incorrectly labeled for a particular class.
    \item \textbf{False Negative (\(FN\))}: denotes an unlabeled pixel that should have been assigned to the class under evaluation.
    \item \textbf{Inference Time (\(L\))}: This is defined as the time taken by a model to process a tensor input and generate an output tensor.
    \item \textbf{Number of Images (\(N\))}: This represents the total number of images comprising the test data set.
\end{enumerate}

\begin{gather}
    \label{equation:dice}
    \begin{split}
        {\small \text{Mean Dice Score Coefficient}} &= \frac{1}{n}\sum_{i=1}^{n}\frac{1}{|C_i|}X \\
    \end{split} \\
    X = \sum_{j=1}^{C_i}\frac{2 \cdot TP_{ij}}{2 \cdot TP_{ij} + FN_{ij} + FP_{ij}} \nonumber
\end{gather}

To evaluate the accuracy of a model, we decided to use the Dice Score Coefficient (DSC), an accepted metric for measuring the similarity between segmentation maps. In Equation (\ref{equation:dice}) we calculate the mean Dice Score Coefficient (mDSC) for each image $i$, by evaluating the Dice Score Coefficient for each class in set $C_i$. The mDSC was then averaged across all images in the test data set, resulting in the accuracy score for a submitted model.

The decision to use mDSC as the accuracy calculation was driven by the significant consequences it exacts for False Positives (\(FP\)) or False Negatives (\(FN\)). To illustrate, consider a scenario where the ground truth, set \(C_g\), contains three classes, while the set of classes in the prediction map \(C_p\) comprises of four classes. In such a case of predicting an extra class, the accuracy of that prediction map would suffer a significant loss. This loss arises because the mDSC becomes an average over four classes rather than the original three classes in the ground truth, per the definition of \(C\) and Equation (\ref{equation:Classes}). The additional class, having no True Positives (\(TP\)), would yield a DSC score of 0. Consequently, if the three classes were otherwise near perfectly predicted, the mDSC would drop from approximately 1.0, amongst the three classes in the ground truth, to 0.75 with the inclusion of the incorrectly predicted class.

The choice of a strict calculation metric is rooted in the nature of the competition, which demands that UAV-captured segmentation maps accurately represent the classes present in an image. For instance, if a UAV incorrectly labeled a person in the middle of a flood, it may trigger a search and rescue mission even though the person class was falsely predicted in the segmentation. Conversely, failing to detect the presence of a person class in the image should maintain a significant penalty to uphold the accuracy of such critical determinations.

To measure the efficiency of a model we calculated its mean inference time using Equation (\ref{equation:time}).
\begin{equation}
\label{equation:time}
\text{Inference Time} = \frac {\sum_{i=1}^{n}L_i}{n}
\end{equation}

Further, we introduce the scoring metric in Equation (\ref{equation:score}), which represents the ratio of accuracy to inference time. This score was used to encapsulate the trade-off between the evaluation metrics.
\begin{equation}
\label{equation:score}  
\text{Score} = \frac {\text{Accuracy}}{\text{Inference Time}}
\end{equation}

In this way, our evaluation metrics provide a comprehensive assessment that balances computational efficiency and accuracy, catering to the unique demands of this competition.

\subsection{Referee System}

\begin{figure}[ht]
    \includegraphics[width=0.48\textwidth]{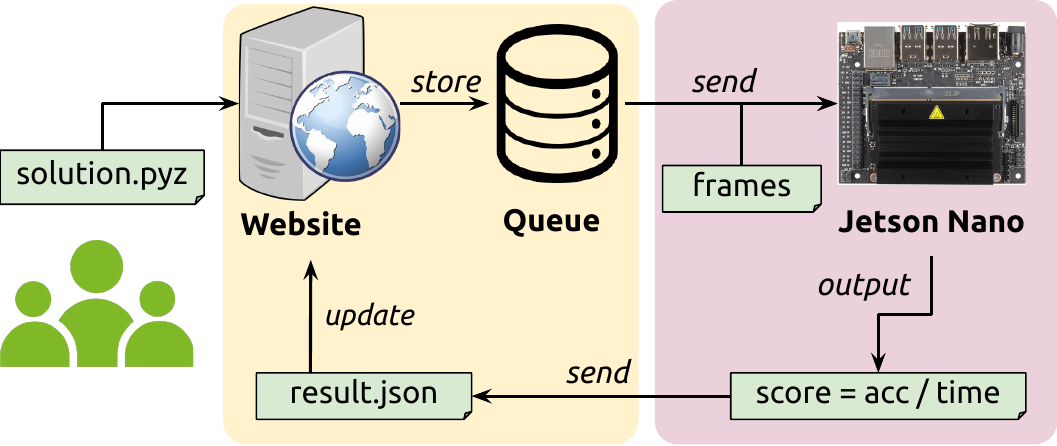}
        \caption{The automated referee system.}
        \label{fig:referee}
\end{figure}

 Figure~\ref{fig:referee} illustrates the information flow in the automated referee system. First, a competitor submits their model and supporting files in a zip file to the competition website for evaluation ({\tt https://lpcv.ai}). The submissions are then queued for evaluation within the referee system. During evaluation, the zip file is sent to the Jetson Nano. The submission yields a cumulative model inference time for processing the 600 test images and the mean inference time is calculated. The referee system subsequently calculates the mDSC across all prediction maps. These metric calculations as well as the performance score are then sent back to the web server, where the submission scores are then posted to the website's leaderboard. A submitted solution is disqualified if it is inferior to the Reference Solution in any metric.

 A submitted solution receives two inputs: the file location to the test images and a file location to the designated output directory. The expected output of a solution is a 512x512 prediction map as well as printing the total summation of the model inference time. A submission is disqualified if it does not produce the correct amount of output images, or if the submission fails to run successfully.

\section{Winners' Solutions}

\begin{table}[ht]
  \small 
  \centering
  \resizebox{\linewidth}{!}{%
    \begin{tabular}{@{}lrrrrl@{}}
      \toprule
      Team       & Accuracy & Time (ms) & Score   & Count  \\ \midrule
      ModelTC    & 0.51   & 6.8    & 75.6 & 33               \\
      AidgetRock & 0.55   & 15.1    & 36.7 & 20                   \\
      ENOT       & 0.60   & 67.0    & 9.0  & 23                    \\ \midrule
      Reference  & 0.50   & 108.1   & 4.6  & 1                 \\
      \bottomrule
    \end{tabular}%
  }
  \caption{Scores of the winning teams compared with sample solution. The last column shows
  the number of submissions from each team.}
\end{table}

\subsection{Team ModelTC}   
    The first-place team, ModelTC achieved a score of 75.608, an accuracy of 51.2\% mDSC, and an average inference time of 6.8 ms. They focused on maximizing inference efficiency while maintaining an accuracy above the sample solution. Their model named the Yocto-Revival Network, is based on a simple UNet. By applying dynamic network techniques~\cite{han2021dynamic}, their model achieved further reduction in inference time~\cite{han2023latency, han2022learning, han2023dynamic, han2022latency, yang2020resolution, zheng2023dynamic, wang2021not, huang2022glance, wang2021adaptive} and improved performance in downstream tasks including segmentation~\cite{pu2023adaptive, pu2023rank, yang2023adaptive}.

    Before settling on their final model, Team ModelTC experimented with PSPNet~\cite{zhao2017pyramid}. The model's quantization was promising, reducing the feature map size significantly. However, the small feature map caused severe drops in accuracy. As a result, they adopted the simple yet effective UNet as the base model.   

    The team observed the most significant breakthroughs in the following modifications. The first was adopting the re-parameterization technique for both training and inference. This approach led the model to achieve higher accuracy while maintaining inference efficiency. Additionally, ModelTC observed that optimizing batch size led to an increase in GPU frequency, thus speeding up inference time.  Moving forward, the team is considering incorporating techniques from a paper on fine-grain recognition~\cite{pu2023fine} to further enhance their model's performance.  
    \url{https://github.com/ModelTC/LPCV_2023_solution}
    
\subsection{Team AidgetRock}
    Team AidgetRock was the winner of the speed award, with a winning accuracy of 55.4\% mDSC, a time of 15ms, and a score of 36.9. Their final model is based on TopFormer~\cite{zhang2022topformer} due to the following features. First, the ability to scale in regards to the size of the features. In the competition dataset, targets may vary significantly in size, for example, a person is small while lava flow typically takes up many pixels. TopFormer uses a feature pyramid network, which proved effective for this task~\cite{yan2022fully, yan2023transy, wan2022siamese}. Second, the migitation of issues caused by poor image quality. Aerial images are often of high resolution and vary in image conditions. To resolve this issue, TopFormer uses Transformers as a backbone and applies an attention mechanism to enhance the concatenated tokens from multiple levels, both of which are known to be powerful in modeling long-range contextual information~\cite{yan2022fully,yan2023transy}. Compared to other models, it is lightweight and performs well in accuracy in the provided datasets. They further optimized the model with pruning~\cite{liu2020autocompress} to reduce the computation cost and apply knowledge distillation (KD)~\cite{zhu2023scalekd} to improve the accuracy of slim models.

    They also attempted to use other models such as SegFormer~\cite{xie2021segformer} and SeaFormer~\cite{wan2023seaformer}, which both are reported to be top-notch in accuracy and speed. However, they found that SegFormer was inefficient while SeaFormer was too large to be trained efficiently. In addition, the accuracy of the models was only comparable to the sample solution, which they believed was caused by overfitting.
    
    They experimented with multiple input sizes and found that pruning led to less than a 1\% drop in dice coefficient accuracy and more than 10\% increase in speed, while knowledge distillation could boost the accuracy of the pruned model by 2\% at the best case. They relied on data augmentation instead of addition due to additional data leading to drops in performance. Furthermore, they used the classic techniques of resizing, cropping, and flipping as well as Mixup and Mosaic augmentation.

    Team AidgetRock adhered to standard pre-processing methods used in the majority of MMSegmentation frameworks~\cite{contributors2020mmsegmentation}. For post-processing, they first converted the trained weight into ONNX format, performing graph optimizations such as removing the Softmax operator and discarding unused blocks. Next, the weights were converted to TensorRT for GPU acceleration. It is worth noting that before removing the Softmax operator the converting failed on the NVIDIA Jetson Nano platform due to a memory error, although the reason for this error was not discovered.
        
    Throughout the process, the team noticed that minor manually labeled data led to some improvements in accuracy. They believed it was due to the quality of the annotated data and the long-tail distribution of the number of pixels in different classes. Dice loss was utilized to mitigate the effect, but the improvement seemed to be limited.

    Team AidgetRock's biggest breakthrough was using ImageNet to pre-train the modified TopFormer backbone. At first, they loaded the original TopFormer weights to reduce the number of layers and blocks. This caused a drop in accuracy, failing to reach the cutoff requirement. As a result, they pre-trained the backbone in the image classification task and initialized the segmentation network with the optimized weight. This led to an improvement of more than 10\% in accuracy. 

    \url{https://github.com/midea-ai/LPCVC2023-AidgetRock}
    
\subsection{Team ENOT}
    
    Team ENOT was the winner of the accuracy award. Their winning submission had a score of 8.974, an accuracy of 60.1\% mDSC, and an average inference time of 67ms. The model in their final solution is based on the PIDNet~\cite{xu2023pidnet}. The architecture was inspired by PID controllers. It also has three branches: P – for details preservation, I – for context embeddings, D – for boundary detection. The main difference from the original PIDNet is the final feature map upsampling method. They replaced the last convolution operation with transposed convolution, improving across the board. They believed it to be an effective model due to the target device and runtime framework (TensorRT). They further optimized the model by removing softmax operation in the end, because they needed argmax from predictions and did not need confidence for each class. After the competition, they theorized that a change in input size would increase model efficiency. In comparison with other teams, ENOT used additional samples from the UAVid~\cite{LYU2020108} dataset. UAVid classes were mapped to the LPCV-2023 classes.

    The biggest breakthrough ENOT experienced was the resize replacement to transposed convolution. They believed this made such a big change due to two reasons: The nearest neighbor resize operation is slow on the Jetson Nano’s GPU and transposed convolution is faster. Transposed convolution is trainable, however, resize is not, which is why the model accuracy only increased after such a replacement. This led to an improvement of +6\% accuracy.
    
    The target metric prevents the prediction of external classes by design, so ENOT used some heuristics to clean up the predictions. They understood that “cracks” were rare and didn't appear concurrently with “lava flow”, so they replaced the “lava flow” with “background” if “cracks” also appeared in the same image. To avoid noisy predictions they replaced objects with less than 500 total pixels in the image to “background”.

    In development, ENOT experimented with DDRNet~\cite{DDRNet_Pan}, Seaformer~\cite{wan2023seaformer} and UNet. DDRNet is a real-time semantic segmentation architecture, which includes two branches with high and low resolution feature maps. They thought that the high-resolution branch addition would help the network perform fine-grained segmentation. Simiarly to AidgetRock, they attempted to use SeaFormer because it was built for mobile semantic segmentation. ENOT believed that the architecture needs more memory to be efficient on the edge devices. UNet was tried as a popular baseline architecture, which could be improved by upsample replacement (from resize to transposed convolution), pruning and knowledge distillation. However, PIDNet performed better in terms of accuracy, so ENOT selected it as a baseline.

    \url{https://github.com/ENOT-AutoDL/lpcv-2023}
    
\subsection{Commonalities of the teams} 
    There are several common features seen in the winning teams: they performed optimizations that improved their models beyond the other teams that who did not make similar improvements. 

    In a competition based on low-power edge devices, improving the utilization of the limited hardware resources is essential. Since TensorRT emerged, it has proved itself superior to previous models ~\cite{TensorRT} due to its ability to maximize GPU usage. The framework excels in model inference time, which it achieves with the following optimizations: 

    \begin{enumerate}
        \item 
    Precision: maximizes throughput by quantizing model to 8-bit integer/16-float integer. 

    \item 
    Fusion: by fusing nodes in a kernel vertically or horizontally (or both), overhead and the cost of writing/reading memory is reduced. 
    
    \item Auto-tuning: provides kernel-specific optimization which selects the best layers, algorithms, and optimal batch size based on the target GPU platform. 
    
    \item Dynamic Tensor Memory: improves memory usage by allocating memory to the tensor only for the duration of its usage. This helps in reducing memory consumption and avoiding allocation overhead for efficient execution. 
    
    \item Multi-Thread Execution: processes multiple input streams in parallel. 
    
    \item Time Fusion: optimizes recurrent neural networks (RNNs) over time steps with the dynamically generated kernel. It must be noted that TensorRT requires a Nvidia GPU, Ubuntu, and CUDA, limiting its applicability. 

\end{enumerate}

    It should be noted that multiple teams reported issues with the softmax operator conflicting with TensorRT, but the cause was not determined.
    
    The execution of quantization should also be noted. AidgetRock in particular featured an extensive method for quantization to preserve accuracy while maximizing speed. Their token pyramid features semantic extractors, injection modules, convolution blocks, transformer blocks, and more. The final size of the model after pre-processing for each team was determined with extensive trials. 
    
    A common decision of the winning teams was to use data augmentation over sourcing additional data. Teams tried using crawlers, public databases, etc with little to no success. Other than ENOT, teams cited losing accuracy and speed when using outside data, thus leading them to focus on data augmentations. Most teams used the standard techniques, such as transformations, color modifications, and filters, to pre-process images. 
    
    Another technique that teams used to boost accuracy in post-processing was processing based on ``common sense''. For example, AidgetRock realized that it was unlikely that ``ice'' could appear with ``lava flow'', and replaced ice with ``background'' if both appeared in the same image. In a similar vein, ENOT noticed that ``cracks'' could not appear with ``lava flow'', and so replaced ``lava flow'' with ``background'' if they appeared simultaneously. 
\section{Future Competitions}

Organizers of future competitions may consider
the following factors:

    Smooth communication between the teams and the organizer is necessary:
    \begin{enumerate}
        \item    
    An open channel of communication between organizers and participants (for example: Slack). Our Slack channel saw upwards of 1000 messages exchanged between the organizers and the participants for clarification. In addition, it encouraged participants to assist one another, forging connections that would've otherwise not been formed. 
    
    \item A highly detailed introduction that uses precise language to set the rules of the competition. 
    One team 
    suffered a decrease in performance due to a misunderstanding that the model could only take inputs of 512x512 images and thus did not opt for model quantization. In this field, attention to detail is critical, and a FAQ should be set up for predicted questions as well as common questions that appear throughout the competition.
    
    \item A leaderboard that encourages competition and serves as near real-time response to submissions. 
    
    \item System tests before the start of the competition. Our competition hosted a system test two weeks before the start of the competition, allowing us to test if our system would work with other solutions and allowing participants to familiarize themselves with the system.

\end{enumerate}

    An issue with this competition 
    is the restricting nature of the possible labels for pixels. Our competition only permitted 14 classes for labels, which is not reflective of realistic disaster scenes. This was one possible reason why two of the winning teams saw a loss of accuracy with data addition. 

    Another restricting feature is the acceptable frameworks that can run on the Jetson Nano. As organizers, it is impossible to support and test every library, every framework, and every GPU optimization technique. For future competitions, we intend to look into submission methods that can mimic the competitor's testing environments. 

    It is natural for there to be unpredictable circumstances that can force an organizer to adapt. For example, due to a power outage, the submission server and evaluation system were down for several days. We added four days to the original month-long competition. Organizers must be prepared to make immediate and effective adaptions to such situations. The aforementioned system test is crucial in detecting errors before they can affect the competition.

\section{CONCLUSION}
This paper reports and analyzes the results of the 2023 Low Power Computer Vision Challenge for semantic segmentation. Since 2015, LPCVC has been growing steadily, with
2023 seeing the largest number of teams
and submissions yet. The winning teams showcased a variety of models, collectively demonstrating the depth of the field. ModelTC developed the Yocto-Revival Network, AidgetRock utilized a recent transformer model called TopFormer~\cite{zhang2022topformer}, and ENOT employed  PIDnet~\cite{xu2023pidnet}, a modern model based on PID controllers.

We hope this paper educates and inspires those who will create the next pivotal computer vision model for low power. In addition, we hope that it serves as a guideline for future competitions of this nature, so that they may advance further growth in this field.

\section{ACKNOWLEDGMENTS}

The organizers want to thank all participants,
especially the winning team 
 ModelTC, AidgetRock, and ENOT for sharing the insight of their solutions.
This project is supported in part by the IEEE Computer Society, IEEE Council on Electronic Design Automation, ACM Special Interest Group on Design Automation, National Science Foundation
OAC 2107230, OAC 2104709, OAC 2104319, OAC 2107020
CNS 2120430. Any opinions, findings, and conclusions or recommendations expressed in this material are those of the authors and do not necessarily reflect the views of the sponsors.

\bibliographystyle{unsrt}
\bibliography{references}

\end{document}